\documentclass{article}

\usepackage{PRIMEarxiv}

\usepackage[utf8]{inputenc} 
\usepackage[T1]{fontenc}    
\usepackage{hyperref}       
\usepackage{url}            
\usepackage{booktabs}       
\usepackage{amsfonts}       
\usepackage{nicefrac}       
\usepackage{microtype}      
\usepackage{lipsum}
\usepackage{fancyhdr}       
\usepackage{graphicx}       
\graphicspath{{media/}}     
\usepackage{amsmath}
\usepackage{enumitem}
\usepackage{makecell}
\usepackage[most]{tcolorbox}
\usepackage{listings}
\usepackage{inconsolata}
\usepackage{xcolor}
\usepackage{multirow}
\tcbset{
    examplebox/.style={
        colback=white,
        colframe=black,
        fonttitle=\bfseries,
        title={#1},
        boxrule=0.5pt,
        arc=1pt,
    }
}

\title{Large Language Models for Medical OSCE Assessment: A Novel Approach to Transcript Analysis
}

\author{
  \normalfont{Ameer Hamza Shakur, PhD$^1$, Michael J. Holcomb, MS$^1$, David Hein, MS$^1$, Shinyoung Kang$^1$,} \\
  Thomas O. Dalton, MD$^2$, Krystle K. Campbell, DHA$^3$, 
  Daniel J. Scott, MD$^{3,4}$, \\ Andrew R. Jamieson, PhD$^1$ \\
  \textit{$^1$Lyda Hill Department of Bioinformatics, $^2$Department of Internal Medicine,} \\
  \textit{$^3$Simulation Center, $^4$Department of Surgery,} \\
  UT Southwestern Medical Center \\
  Dallas \\
  \texttt{\{ameerhamza.shakur, andrew.jamieson\}@utsouthwestern.edu}
  \And
}

\begin{document}
\maketitle

\begin{abstract}
Grading Objective Structured Clinical Examinations (OSCEs) is a time-consuming and expensive process, traditionally requiring extensive manual effort from human experts. In this study, we explore the potential of Large Language Models (LLMs) to assess skills related to medical student communication. We analyzed 2,027 video-recorded OSCE examinations from the University of Texas Southwestern Medical Center (UTSW), spanning four years (2019-2022) and several different medical cases or ``stations''. Specifically, our focus was on evaluating students' ability to summarize patients' medical history \textemdash we targeted the question ‘did the student summarize the patients’ medical history?’ from the communication skills rubric. After transcribing speech audio captured by OSCE videos using Whisper-v3,  we studied the performance of various LLM-based approaches for grading students on this summarization task based on their examination transcripts. Using various frontier-level open-source and proprietary LLMs, we evaluated different techniques such as zero-shot chain-of-thought prompting, retrieval augmented generation, and multi-model ensemble methods. Our results show that frontier LLM models like GPT-4 achieved remarkable alignment with human graders, demonstrating a Cohen's kappa agreement of 0.88 and indicating strong potential for LLM-based OSCE grading to augment the current grading process. Open-source models also showed promising results, suggesting potential for widespread, cost-effective deployment. Further, we present a failure analysis identifying conditions where LLM grading may be less reliable in this context, and recommend best practices for deploying LLMs in medical education settings. 
\end{abstract}


\section{Introduction}
The Objective Structured Clinical Examinations (OSCE) \cite{harden1988osce} is a crucial component of the professional education and training of students during medical school. The OSCE tests various components of clinical competency by putting the student through mock medical encounters with a human actor known as a standard patient (SP), where they role-play as patients suffering from common medical conditions. Several aspects of performance are evaluated during the OSCE, such as students’ proficiency in taking medical history, clinical reasoning, physical examinations, and ability to come to the correct diagnosis and workup. Equally important is testing various dimensions of the student’s communication skills \cite{comert2016assessing} ---  building trust,  giving and providing information, and demonstrating empathy. For each case the standardized patient reports a chief complaint and is trained to respond to student questions on the key attributes of their present illness, associated symptoms, medical history, and family and social history. Then, they perform physical examinations if needed and discuss differential diagnosis and an appropriate treatment plan. The entire encounter is video recorded to be later reviewed by a team of trained experts. 

Encounter videos contain rich information about a student’s ability to perform in a clinical setting. These recordings offer valuable opportunities for students to receive actionable feedback, enhancing their skills and better preparing them for evaluations with actual patients. However, the unstructured nature and length of these videos make thorough assessment challenging. Traditionally, evaluation involves a team of highly trained professional experts watching each video in its entirety and grading students based on specific rubrics. This approach is time-consuming, expensive, and potentially subject to human bias. Furthermore, the lengthy review process often delays feedback delivery to students by weeks, significantly reducing its effectiveness. The time lag between the examination and receiving feedback diminishes the potential for timely improvement in students' clinical skills. 

Advances in automatic speech recognition (ASR) and natural language processing (NLP) have unlocked opportunities for more nuanced, timely, and cost-efficient evaluation of these medical encounters. ASR systems are now approaching human-like proficiency in accurately transcribing conversations  \cite{radford2023robust}. Concurrently, large language models (LLMs) have demonstrated impressive capabilities in tasks such as sentiment analysis, text comprehension, summarization, and even reasoning \cite{singhal2023large}. These technological advancements present an opportunity to reimagine OSCE grading by augmenting it with AI, potentially developing a system that is faster, more economical, and less prone to bias than the status quo.

\subsection*{Prior Work} \label{subsec:lit_rev}
Recent advancements in LLMs have generated significant interest in re-imagining medical education \cite{chan2019applications, masters2019artificial, abd2023large}. Frontier models have demonstrated impressive capabilities in medical domain tasks, with ChatGPT, Med-PaLM, and MedPaLM-2 achieving passing marks in standardized medical entrance examinations such as the USMLE. They have also matched, if not exceeded human performance on several benchmarks measuring knowledge and reasoning capabilities both in a general sense \cite{achiam2023gpt, anil2023palm} as well as specifically in medical Q\&A \cite{singhal2023large, karabacak2023embracing}. The growing capabilities of AI systems have sparked broad discourse about this technology's potential to revolutionize all aspects of medical education --- curriculum development, teaching, learning support, assessment, personalized education, etc. \cite{chan2019applications}. 

Recent efforts to automate OSCE assessment include use of large multi-modal models (LMMs) to evaluate the physical exam component \cite{holcomb2024zero} of the OSCE. Here, conventional computer vision techniques were used to extract relevant frames; then the LMM was prompted to describe and grade the student based on the extracted frames \cite{holcomb2024zero}. A recent work \cite{vedovato2024towards} proposed a deep learning pipeline to identify, extract, and score ear exams based on off-the-shelf object detection tools such as CLIP \cite{radford2021learning} and Detectron2 \cite{wu2019detectron2}. Our literature search found only one study focused on grading communication skills from transcript analysis. In this study, \cite{jani2020machine} 121 OSCE videos were manually transcribed and annotated, and five different text classification models were evaluated on their ability to label various communication skill items from the transcripts.

Our work investigates the feasibility of state-of-the-art LLMs to assess student performance, specifically in history summarization, by analyzing transcripts extracted from the recorded exams.  We measure model alignment with human expert graders from a collection of historical exams made up of 2,027 video-recorded OSCEs from the University of Texas Southwestern Medical Center (UTSW), spanning four years and ten different medical cases or `stations'. Additionally, we  investigate the impact of retrieval augmented generation (RAG) techniques and  provide a comprehensive overview of the various failure modes and considerations for LLMs and RAG-techniques that should be considered prior to deploying such tools for exam grading. We believe this to be the first work to evaluate a transcript-based AI-augmented OSCE grading system at this scale.

\section{Methods}
\subsection{Datasets and pre-processing} \label{sec:data}

We selected 2,027 videos of the Comprehensive Objective Structured Clinical Exams (COSCEs) recorded at UT Southwestern Medical Center (UTSW) between the years 2019 – 2022. Videos covered simulated consultations with patient actors presenting different ‘cases’ (also known as ‘stations’) such as ‘itchy eyes’, ‘vision problems’, ‘memory problems’ etc. The exams are evaluated by a team of trained human experts using standardized checklists or rating scales to assess each candidate’s performance. Table 1 shows the distribution of the counts and scores in the COSCEs by year in our evaluation dataset.

\subsubsection*{Data safety}
Strict data protocols were used to ensure data safety. Proprietary models were accessed via a secure platform. Where this was not available, presidio-anonymizer \cite{presidio-anonymizer} package was used to remove all personally identifiable information from the transcripts. For open-source models that could be run on premise, our secure institutional high performance computing environment, BioHPC, was used.

\begin{figure}[htbp]
    \centering
    \begin{minipage}{0.45\textwidth}
        \centering
        \begin{tabular}{lcc}
        \hline
        \textbf{Year} & \textbf{Count} & \textbf{Received full credit} \\
        \hline
        2019 & 590 & 88.8\%\\
        2020 & 317 & 70.2\%\\
        2021 & 470 & 70.1\%\\
        2022 & 650 & 67.3\%\\
        \hline
        \end{tabular}
        \caption{No. of COSCE exams and proportion of students that received full credit on `summary of medical history' for each year}
        \label{tab:annualcount}
    \end{minipage}
    \hfill
    \begin{minipage}{0.45\textwidth}
        \centering
        \includegraphics[width=0.3\textwidth]{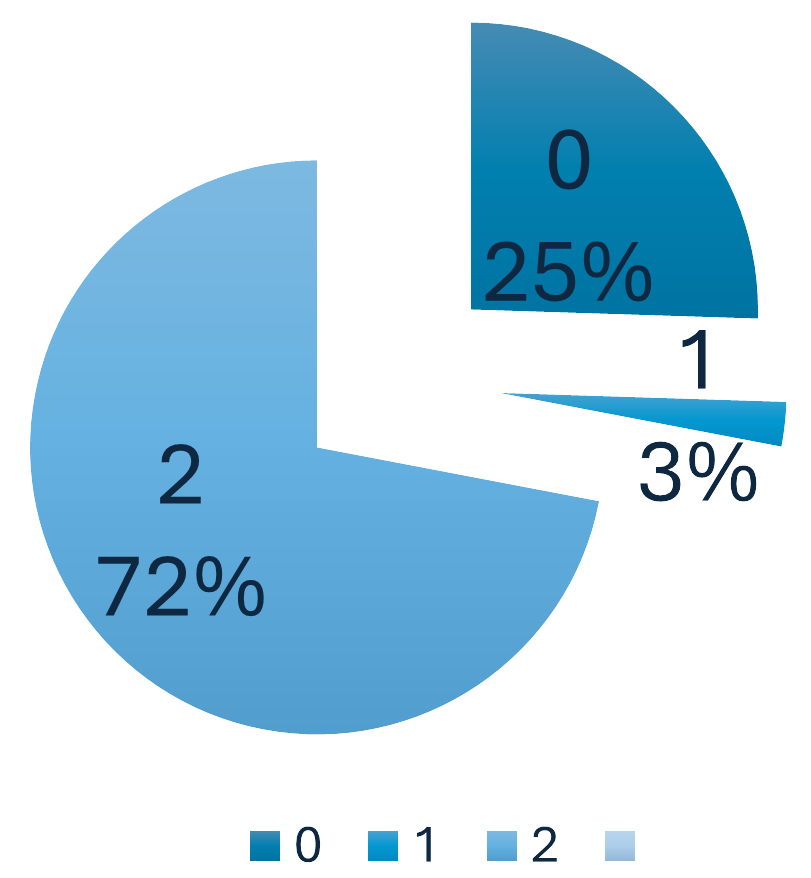}
        \caption{Distribution of scores on `summary of medical history' item of the communication skills rubric, with 2 being full credit, 1 partial credit, and 0 no credit.}
        \label{fig:score_pie}
    \end{minipage}
\end{figure}

\subsubsection*{Summary of medical history task} In this study, we specifically address the item  - ‘did the student summarize the patients’ medical history’? - from the communication skills rubric of the OSCE. Providing a summary of medical history is a crucial skill that serves a critical purpose in addressing the patient’s concerns. It ensures that the student has understood the patient’s overall health status and allows the patient to issue corrections in case of a misunderstanding. For example, an acceptable summary of medical history provided by the student at the ‘memory problems’ station may look like:

“\textit{So just to kind of summarize over the past seven months, you've been feeling that you have been having some difficulty concentrating, it's been taking you some time to balance your checkbook, and you noticed you just sleep less, you had less of an appetite, and you kind of not found as much interest in things that you used to enjoy. And it's been kind associated with some weight loss and no other pain anywhere. Is that correct?}” 

Fig. \ref{fig:score_pie} shows the distribution of scores the medical students obtained on this question, with 2 being full credit. We noticed that approximately 3\% of students received a score of 1 from the SP graders or partial credit. 

\subsubsection*{Pre-processing}
\begin{itemize}
    \item Transcription. OpenAI’s Whisper-large-v3 \cite{radford2023robust} was used for automated speech recognition algorithm to generate a transcript of the conversation that took place in the examination videos. No further human corrections or annotations were used in the transcript. We found that ~2\% of the transcripts were unusable due to poor audio quality of the exam recordings \textemdash. These were removed from the analysis. 
    \item Anonymization. As mentioned above, if needed, the presidio-anonymizer was also run to remove personally identifiable information before sending it to Claude APIs. 
\end{itemize}

\begin{figure}
    \centering
    \includegraphics[width=0.9\linewidth]{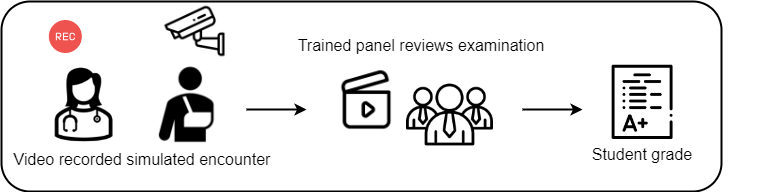}
    \caption{Schematic showing the current human-expert grading process of OSCE's. Cameras in the examination room record the OSCE encounters, which are then reviewed by two trained human experts who grade the student examination.}
    \label{fig:current-schematic}
\end{figure}

\subsection{Model Design} \label{sec:methods}

In this section, we describe the LLMs and different prompting strategies explored in this study as well as our criteria for evaluating system performance against human expert graders. Table \ref{tab:model-comparison} provides an overview of the models evaluated. We selected these models to represent a broad spectrum of current LLM capabilities, including  state-of-the-art proprietary models and capable open-source alternatives. Two primary approaches for automated grading of OSCE’s were investigated – zero-shot grading and retrieval-augmented generation.

\begin{table}[htbp]
\centering
\caption{Large Language Models Used for Transcript Analysis}
\begin{tabular}{p{4cm}p{3cm}p{2cm}p{3cm}}
\hline
\textbf{Model Name} & \textbf{Developer} & \textbf{Type} & \textbf{Access} \\
\hline
GPT-4 (01-25) & OpenAI & Proprietary & Azure OpenAI API \\
GPT-4o (02-15) & OpenAI & Proprietary & Azure OpenAI API \\
GPT-3.5 (06-13) & OpenAI & Proprietary & Azure OpenAI API \\
Claude-Opus (04-22) & Anthropic & Proprietary & Anthropic API \\
Claude-Sonnet-3-5 (06-27) & Anthropic & Proprietary & Anthropic API \\
Llama-3-70B & Meta AI & Open-source & Local deployment \\
Llama-3-8B & Meta AI & Open-source & Local deployment \\
Mixtral 8x7B & Mistral AI & Open-source & Local deployment \\
Starling-7B-beta & Berkeley AI Research & Open-source & Local deployment \\
\hline
\end{tabular}
\label{tab:model-comparison}
\end{table}

\begin{figure}[!htpb]
    \centering
    \includegraphics[width=0.9\linewidth]{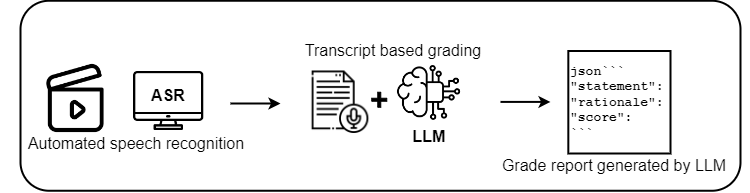}
    \caption{Schematic representation of our zero-shot grading workflow for OSCE assessments. First, Whisper-v3 speech recognition is used to transcribe the recorded OSCE encounters. Then the transcript is analyzed by an LLM to assess student performance.}
    \label{fig:zero-shot-schematic}
\end{figure}

\subsection{Zero-shot generation} Zero-shot generation leverages the LLM's knowledge and "reasoning" ability without any explicit training or learning from prior OSCE transcripts or graded examples. In this approach, we provide the entire transcript to the model along with the grading instructions for the rubric item and a scoring criterion. Further, we leveraged chain-of-thought prompting \cite{wei2022chain}, which has been shown to improve LLM reasoning ability by asking the model to articulate internal reasoning steps before arriving at the grade. In our case – we prompt the model to first extract the relevant ‘statement’ which is the portion of the transcript that corresponds to the item being graded, then generate a ‘rationale’ based on the grading rubric and a ‘score’ based on the scoring criteria.  Finally, we ask the model to return each of these items in a structured JSON format for convenient post processing. Fig. \ref{fig:zero-shot-schematic} illustrates our workflow for the zero-shot grading process. Prompt instructions are provided in the supplementary materials.

\begin{figure}[!htpb]
    \centering
    \includegraphics[width=0.9\linewidth]{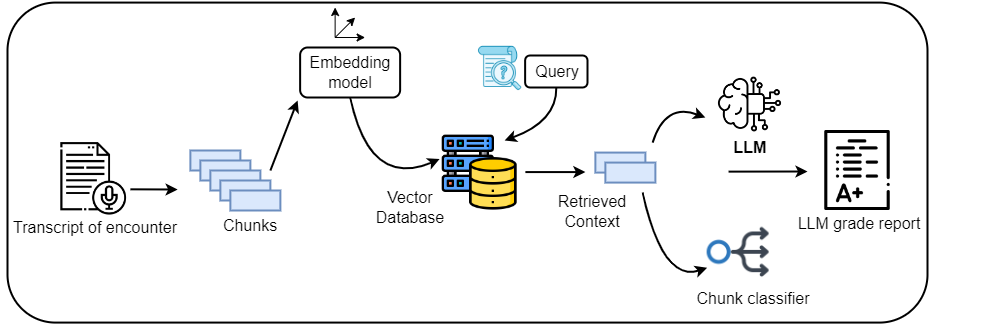}
    \caption{Schematic of retrieval-augmented grading workflow for OSCE exams, combining LLM capabilities with embedding-based information retrieval.}
    \label{fig:rag-shematic}
\end{figure}

\subsection{Retrieval augmented generation (RAG)} 
Retrieval augmented generation has become a popular technique for enhancing model performance by combining LLM capabilities with information retrieval  \cite{li2022survey, gao2023retrieval}. The primary motivation for using RAG is the ability to produce more accurate, factual, and contextually relevant outputs by grounding the model's responses in retrieved information. Grounding responses with RAG mitigates hallucinations common in language model outputs and enables the handling of specialized tasks without extensive fine-tuning. In our workflow, the OSCE transcript is of a 15-minute-long conversation and only a small portion of it is relevant to the rubric item being graded. By retrieving only the most relevant portion of the transcript for the model to grade, performance may be improved. Additionally, fewer tokens will be required, reducing inference costs. The retrieval augmented grading process involves two primary steps: i) retrieving the most relevant sections of the transcript in the current context and ii) prompting the LLM to generate a score based on the retrieved contexts. We describe our approach for each of these steps in further detail.

\begin{table}[h]
\centering
\begin{tabular}{p{0.2\textwidth}p{0.7\textwidth}}
\hline
\textbf{Query strategy} & \textbf{Query explanation} \\
\hline
Baseline & The statement `a summary of medical history of a patient presenting with the following case: <\textit{case}>' \\
HyDE (Hypothetical document embeddings) & An LLM generated summary of medical history of a patient suffering from <\textit{case}> spoken from a doctor's perspective \\
Script-based & An LLM generated summary of medical history given the script of the examination encounter \\
Auto-summarizer & An LLM generated summary of the medical history reported by the patient given the transcript of the encounter \\
\hline
\end{tabular}
\caption{Explanation of query strategies used for embedding-based retrieval of medical history summaries. Full prompt examples used to generate queries are provided in the Supplementary Materials [\ref{sec:prompts}].}
\label{tab:query_strategies_methods}
\end{table}

\subsubsection{Retrieving the correct context}
The retrieval of the relevant portion of the transcript involves the following sequential steps:
\begin{itemize}
    \item \textbf{Chunking}: We segment the full transcript into an average of 25 sequential chunks. We used the Lang Chain \cite{langchain} text-splitter with a chunk length of 500 characters and overlap of 0 characters. These parameters were chosen to optimize retrieval using a validation dataset.
    \item \textbf{Embedding}: We then create vector embeddings of each of the chunks using the \textit{bge-large-v2} embedding model \cite{bge-large-v2}. Each chunk is encoded into a high-dimensional embedding space of 1024 dimensions, where the embedding vector can be understood to preserve some semantic context. These vector embeddings were then stored in a \textit{Chroma} vector database. 
    \item \textbf{Querying}. To retrieve the most relevant chunks of the transcript we use a “query” vector that is semantically similar the chunk that we are looking to retrieve from the transcript. The choice of this query vector is the most crucial step to determining the context being retrieved by the retriever. We experimented with several query formulation strategies, including using predefined templates, dynamically generated queries based on the rubric item, and even using LLMs to generate context-aware queries. The different query strategies used are elucidated in Table \ref{tab:query_strategies_methods}. 
    \item \textbf{Reranking}. The chunks are then re-ranked based on their relevance to the query. We used the \textit{ms-marco-TinyBERT-L-2} cross-encoder model \cite{ms-marco-tinybert}. This model is trained on query-vector pairs and outputs a relevance score for each pair of sentences. 
    \item \textbf{Retrieval}. The top 5 most relevant chunks based on the re-ranker score are then chosen to be retrieved.
\end{itemize}

\subsubsection{Retrieval augmented grading} \label{sec:RAG_models}
We investigate two different approaches for grading students based on the retrieved context from the transcripts:
\begin{itemize}
    \item \textbf{LLM grading}. In this approach, we send the retrieved context from the retrieval step to the LLM with a prompt nearly identical to that used in the zero-shot generation process. The model is then asked to grade the students based solely on this retrieved context.
    \item \textbf{SetFit Classification}. In this approach, we utilize SetFit (Sentence Transformer fine-tuning) \cite{setfit}, an efficient few-shot text classification method. SetFit works by fine-tuning a pre-trained Sentence Transformer embedding model on a small number of pairs of sentences in a contrastive manner to create task-specific embeddings. This is followed by training a classification head on these fine-tuned embeddings. SetFit has been shown to achieve competitive performance with as few as 8 labeled examples per class, outperforming larger language models and being far more computationally efficient. In our implementation, we trained a multi-label SetFit classifier on manually labeled chunks of the transcript to identify whether any constitute a ``summary statement''.
\end{itemize}
By implementing these two distinct approaches, we aimed to compare the effectiveness of a powerful, general-purpose LLM against a more specialized, efficient classifier in the context of OSCE grading. This comparison provides insights into the trade-offs between model complexity, computational requirements, and grading accuracy with RAG.

\subsection{Evaluation criteria}
We employed several evaluation metrics to assess the performance of our RAG and LLM-based grading system for OSCE exams. These metrics were chosen to capture various aspects of the grading process, including accuracy, consistency, and the effectiveness of the retrieval mechanism.

\subsubsection{Evaluating transcript grading performance}

The following metrics were used to evaluate the overall performance of the automated grading system:

\begin{itemize}
    \item \textbf{Accuracy score}: This metric measures the overall correctness of the grading system by calculating the proportion of correctly graded responses out of the total responses. It is defined as:
    
    \begin{equation}
        \text{Accuracy} = \frac{\text{Number of correctly graded responses}}{\text{Total number of responses}}
    \end{equation}
    
    While accuracy provides a straightforward measure of performance, it may not be sufficient in cases of imbalanced datasets.

    \item \textbf{F1 score}: This metric provides a balanced measure of the grading system's precision and recall, making it particularly useful for evaluating performance on imbalanced datasets. The F1 score is the harmonic mean of precision and recall:
    
    \begin{equation}
        \text{F1 score} = 2 \cdot \frac{\text{Precision} \cdot \text{Recall}}{\text{Precision} + \text{Recall}}
    \end{equation}
    
    where Precision is the ratio of true positives to all predicted positives, and Recall is the ratio of true positives to all actual positives.

    \item \textbf{Cohen's kappa}: This statistic assesses the inter-rater agreement between the automated grading system and human graders, accounting for the possibility of agreement occurring by chance \cite{fleiss1973equivalence}. It is calculated as:
    
    \begin{equation}
        \kappa = \frac{p_o - p_e}{1 - p_e}
    \end{equation}
    
    where $p_o$ is the observed agreement and $p_e$ is the expected agreement by chance. Cohen's kappa provides a more robust measure of agreement than simple percent agreement calculation.
\end{itemize}

\subsubsection{Evaluation of retrieval performance}

To assess the effectiveness of the retrieval component in our RAG system, we calculated the recall metric,

\begin{itemize}

    \item \textbf{Retrieval recall}: This metric measures the proportion of relevant information successfully retrieved for grading purposes. It is defined as:
    
    \begin{equation}
        \text{Retrieval Recall} = \frac{\text{Number of retrieved summary statements}}{\text{Total number of transcripts}}
    \end{equation}
    
\end{itemize}

To calculate this metric, we used a separate dataset of exam transcripts where GPT-4's zero-shot grading matched human grading. For these exams, we extracted ``summary statements'' identified by GPT-4 that were manually verified to be accurate. Our retrieval method then produced the top-5 chunks of each transcript. If the corresponding summary statement was found in these chunks, we considered it a successful retrieval. The retrieval performance was measured as the proportion of exams where the summary statement was successfully retrieved within the top-5 chunks. 

\section{Results} \label{sec:results}

\subsection{Alignment of LLM grades with human expert graders}

\begin{figure}
    \centering
    \includegraphics[width=\linewidth]{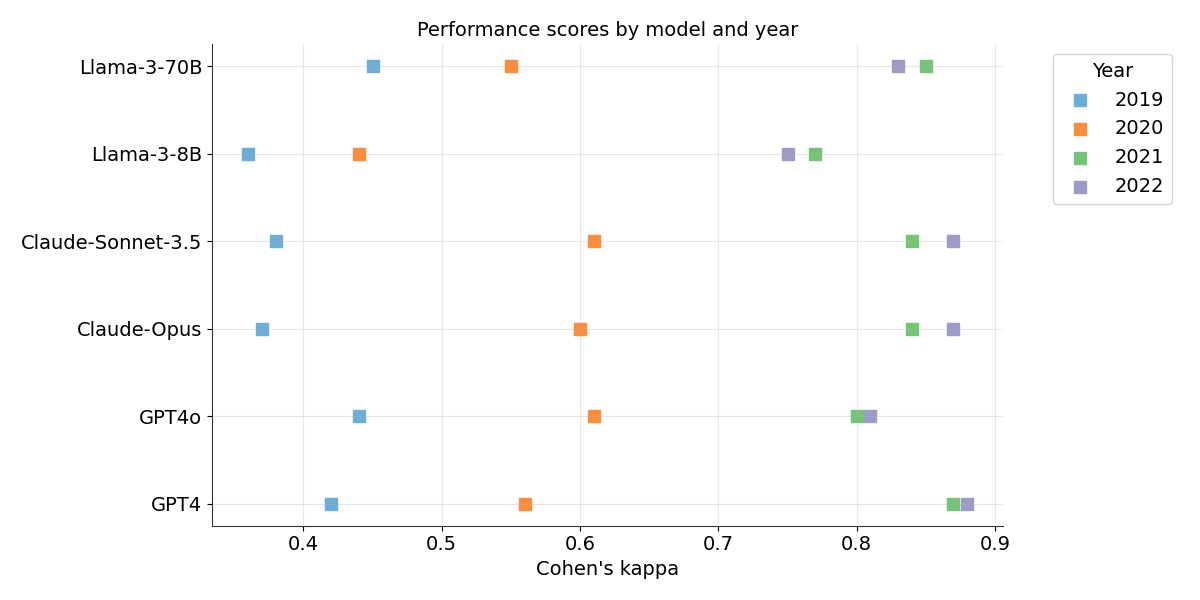}
    \caption{Comparison of model performance in terms of Cohens kappa on grading ``summary of medical history'' across different years}
    \label{fig:yearly_barchart}
\end{figure}
\begin{table}[h]
\centering
\begin{tabular}{lcccccc}
\hline
\textbf{Year} & \textbf{GPT-4} & \textbf{GPT-4o} & \textbf{Claude-Opus} & \textbf{Claude-Sonnet-3-5} & \textbf{Llama-3-8B} & \textbf{Llama-3-70B} \\
\hline
2019 & 0.42 & 0.44 & 0.37 & 0.38 & 0.36 & 0.45 \\
2020 & 0.56 & 0.61 & 0.60 & 0.61 & 0.44 & 0.55 \\
2021 & 0.87 & 0.80 & 0.84 & 0.84 & 0.77 & 0.85 \\
2022 & 0.88 & 0.81 & 0.87 & 0.87 & 0.75 & 0.83 \\
\hline
\end{tabular}
\caption{Performance comparison of language models across four different years}
\label{tab:model_performance_years}
\end{table}

Fig. \ref{fig:yearly_barchart} shows a longitudinal view of LLM and human grade alignment across four distinct student cohorts from 2019 to 2022. A dramatic improvement in alignment was found in the year 2021. GPT -4's performance best exemplifies this shift: its kappa value increased from 0.56 in 2020 to over 0.87 in 2021, further improving to 0.88 in 2022 (Table \ref{tab:model_performance_years}). This improvement was mirrored across other models, suggesting a fundamental change in human grading approach rather than model-specific enhancements.

\begin{figure}[!htpb]
    \centering
    \includegraphics[width=0.95\linewidth]{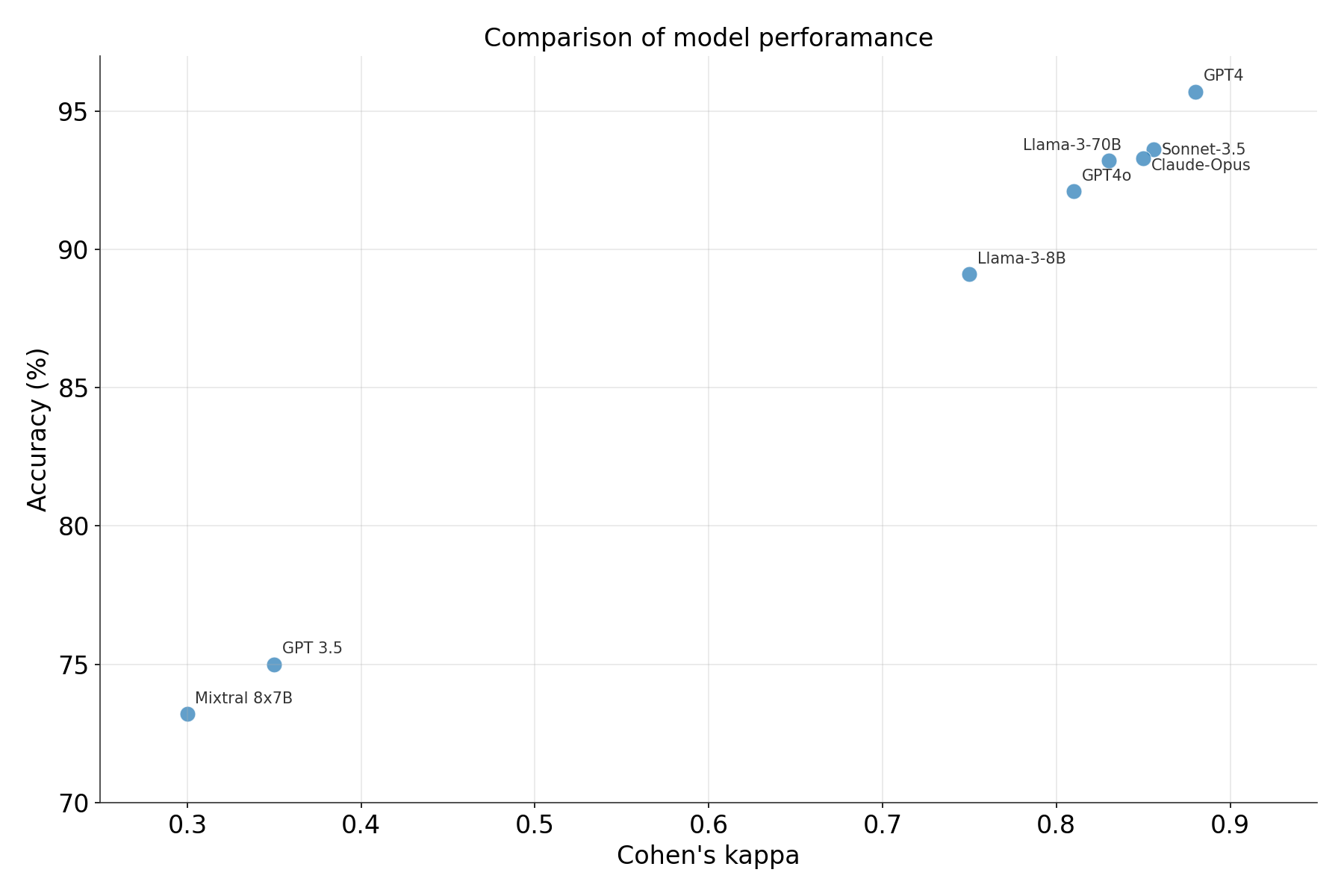}
    \caption{Comparison of model performance in terms of accuracy score and Cohens' kappa vs human graders for years after 2020. Starling-7B is not represented in this plot due to its poor performance being out of bounds.}
    \label{fig:scatter1}
\end{figure}

Therefore, focusing only on data from 2021 onwards, we show the comparative performance of each of the models in Fig. \ref{fig:scatter1}. Here we observe a clear hierarchy in model performance. GPT-4 emerged as the clear front runner, achieving a  95.7\% accuracy and a kappa of 0.88. Next, we find that Claude-Opus and Claude-Sonnet-3-5 and the open-weights Llama-3-70B are closely matched in terms of their performance, achieving Cohens kappa values ranging between 0.86 to 0.83 with Sonnet-3-5 having a slight edge over the others. However, GPT 3.5, Mixtral and Starling-7B-beta perform much worse with a kappa value lower than 0.35 for each of these models. 

We further explore performance across different medical cases in Table \ref{table:performance_cases}. Kappa values generally ranged between 0.7 and 0.9 for top-performing models, with slight fluctuations depending on the specific case. This consistency across diverse medical scenarios demonstrates robustness of these models in handling varied clinical contexts.

\begin{table}[h]
\centering
\begin{tabular}{lccc}
\hline
\textbf{Model} & \textbf{Accuracy (\%)} & \textbf{F1 score} & \textbf{Cohen's kappa} \\
\hline
GPT-4 (01-25) & 95.7 & 0.95 & 0.88 \\
Sonnet-3-5 (06-27) & 93.6 & 0.94 & 0.86\\
Claude-Opus (04-22) & 93.3 & 0.93 & 0.85 \\
Llama-3-70B & 93.2 & 0.93 & 0.83 \\
GPT-4o (05-21) & 92.1 & 0.89 & 0.81 \\
Llama-3-8B & 89.1 & 0.89 & 0.75 \\
GPT 3.5 (06-13) & 75.0 & 0.75 & 0.35 \\
Mixtral 8x7B & 73.2 & 0.73 & 0.30 \\
Starling-7B-beta & 69.0 & 0.65 & 0.14 \\
\hline
\end{tabular}
\caption{Performance comparison of various LLMs compared with human graders for years after 2020}
\label{tab:model_performance}
\end{table}

\begin{table}[!htpb]
\centering
\begin{tabular}{lcccccc|c}
\hline
\textbf{Case} & \textbf{GPT-4} & \textbf{GPT-4o} & \textbf{Claude-Opus} & \textbf{Sonnet-3-5} & \textbf{Llama-3-70B} & \textbf{Llama-3-8B} & \textbf{\textit{Std.}} \\
\hline
Cough & 0.79 & 0.72 & 0.76 & 0.76 & 0.82 & 0.81 & 0.034\\
Itchy eyes & 0.88 & 0.70 & 0.79 & 0.79 & 0.75 & 0.59 & 0.093\\
Jaundice & 0.84 & 0.83 & 0.77 & 0.81 & 0.74 & 0.72 & 0.045\\
Leg pain & 0.89 & 0.86 & 0.91 & 0.91 & 0.90 & 0.75 & 0.056\\
Memory problems & 0.87 & 0.79 & 0.87 & 0.87 & 0.73 & 0.81 & 0.042 \\
Nausea & 0.87 & 0.80 & 0.83 & 0.83 & 0.85 & 0.66 & 0.069\\
No weight gain & 0.97 & 0.76 & 0.94 & 0.94 & 0.85 & 0.75 & 0.034 \\
Vaginal discharge & 0.90 & 0.88 & 0.85 & 0.85 & 0.90 & 0.83 & 0.027\\
Vision problems & 0.82 & 0.75 & 0.82 & 0.81 & 0.88 & 0.78 & 0.037\\
\hline
\textbf{\textit{Std.}} & 0.047 & 0.067 & 0.049 & 0.047 & 0.063 & 0.079 &  \\
\hline
\end{tabular}
\caption{Comparison of model performance on various medical cases for years after 2020}
\label{table:performance_cases}
\end{table}

\subsection{Retrieval augmented generation}

\subsubsection{Retrieval performance}
We first evaluated the effectiveness of different query strategies for retrieving the relevant portions of the transcript containing the summary of medical history. Table \ref{tab:query_strategies_results} presents a comparison of these strategies. When the query-vector is simply the statement "a summary of medical history of a patient presenting with <\textit{case}>", the recall@5 is only 54\%. However, the auto-summarizer strategy emerged as the most effective, achieving a remarkable 92\% recall@5. This means that for 92\% of the transcripts, the relevant summary section was found within the top 5 retrieved chunks. This strategy involves using an LLM to generate a summary of the medical history based on the full transcript, which is then used as the query vector.

\begin{table}[h]
\centering
\begin{tabular}{ccccc}
\hline
\textbf{top-k} & \textbf{Baseline}  & \textbf{HyDE}  & \textbf{Script}  & \textbf{Auto-summarizer} \\
\hline
1 & 27\% & 35\% & 45\% & 63\% \\
2 & 39\% & 48\% & 62\% & 77\% \\
3 & 43\% & 59\% & 73\% & 84\% \\
4 & 49\% & 67\% & 79\% & 90\% \\
5 & 54\% & 72\% & 84\% & 92\% \\
\hline
\end{tabular}
\caption{Comparison of recall @ top 1 - 5 retrieved chunks for different query strategies in medical history retrieval. Strategies are explained in Table \ref{tab:query_strategies_methods}.}
\label{tab:query_strategies_results}
\end{table}

\begin{figure}
    \centering
    \includegraphics[width=0.75\linewidth]{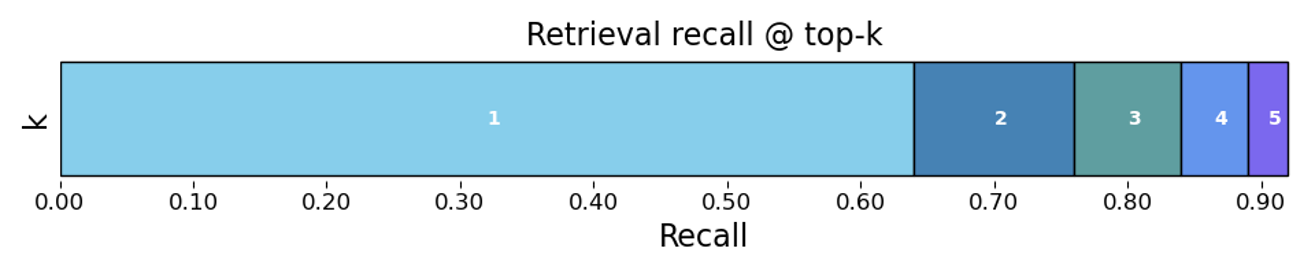}
    \caption{Visual representation of recall performance of retrieval system at top 1 - 5 chunk for the\textit{ auto-summarizer} retrieval strategy}
    \label{fig:retrieval-recall}
\end{figure}

To further illustrate the performance of our best-performing retrieval strategy (auto-summarizer), we examined its recall at different top-k values, as shown in Table \ref{tab:query_strategies_results}. These results demonstrate the rapid improvement in recall as we increase the number of retrieved chunks. Even at top-1, we achieve a respectable 63\% recall (already above the baseline-strategy top-5), rising to 90\% at top-4. This suggests that in most cases, the relevant summary information is concentrated in a small portion of the transcript, supporting the potential for efficiency gains when using a RAG approach. Fig. \ref{fig:retrieval-recall} visually illustrates recall performance as the number of retrieved chunks is increased from 1 to 5. 

\begin{table}[!htpb]
\centering
\begin{tabular}{lcc}
\hline
\textbf{Model} & \textbf{Cohen's kappa} & \textbf{F1 score} \\
\hline
Retrieval + SetFit & 0.62 & 0.84 \\
Retrieval + GPT-4 & 0.56 & 0.82 \\
\hline
\end{tabular}
\caption{Performance comparison of retrieval augmented grading approaches for years after 2020}
\label{tab:retrieval_models}
\end{table}

\subsubsection{Performance of the retrieval augmented grading scheme}
Table \ref{tab:retrieval_models} presents the results of two RAG approaches. The Retrieval + SetFit approach (\ref{sec:RAG_models}) which uses a multi-label SetFit classifier trained on retrieved chunks to identify summary statements achieved a Cohen's kappa of 0.62 and an F1 score of 0.84. Notably, the Retrieval + GPT-4 approach (\ref{sec:RAG_models}), where we sent the retrieved context to GPT-4 for grading, resulted in a Cohen's kappa of 0.56 and an F1 score of 0.82. These results are notably lower than the performance achieved by full transcript grading, where GPT-4 attained a Cohen's kappa of 0.88 (as shown in Table 1). This significant drop in performance is particularly surprising given the high recall rates of our retrieval system. Several factors may be contribute to this performance gap such as loss of context, sensitivity to retrieval errors, or  chunk integration, or needing the full context of the conversation for accurate grading. By focusing only on the retrieved chunks, the models may miss important contextual information present in the full transcript. This could be crucial for accurately identifying and evaluating summary statements. For the LLM approach, the model may struggle to coherently integrate information from multiple retrieved chunks compared to processing a full, continuous transcript. The SetFit classifier may be limited by the quantity and quality of training data available from the retrieved chunks.

\subsection{Inter-model agreement and ensemble-performance}

\begin{figure}[!htpb]
    \centering
    \includegraphics[width=0.85\linewidth]{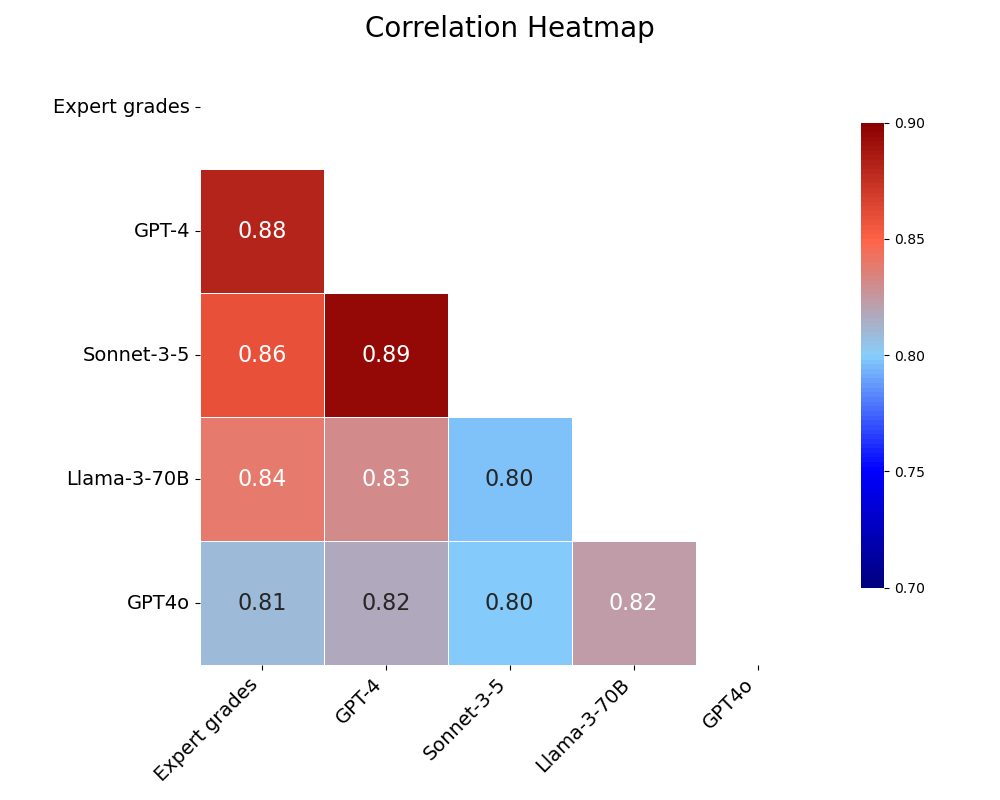}
    \caption{Heatmap of inter-model agreement for OSCE grading representing the pairwise agreement between different language models in grading OSCE examinations for years after 2020.}
    \label{fig:inter_modell}
\end{figure}

We also investigated if grading performance could be improved by using an ensemble of models, for example by offering increased robustness and reliability.  First,  inter-model correlation (Fig. \ref{fig:inter_modell}) can be checked to reveal which models perform most similarily at a high-level.  GPT-4 and Sonnet-3-5 show the highest inter-model agreement at 0.89. Also, we see a pattern of slightly decreasing agreement as we move from top-performing models to lower-performing ones. For instance, GPT-4o shows lower agreement with other models (0.8-0.825) compared to the agreement between top models. The inter model agreement rate also indicates that while models agree on a significant portion of their grades, there's also meaningful variation. If agreement were too low, it would suggest the models are inconsistent or unreliable. If agreement were too high (near 1.0), ensemble methods wouldn't provide much benefit as all models would essentially give the same grades. The slightly different perspectives of each model (as evidenced by agreement levels around 0.8 rather than 1.0) suggest that an ensemble approach could help identify and correct for individual model biases or errors.

\begin{table}[!htpb]
    \centering
    \begin{tabular}{c>{\raggedright\arraybackslash}lcccc}
        \hline
        \textbf{\makecell{\# Models in\\agreement}} & \textbf{Models} & \textbf{\makecell{\% of the\\dataset}} & \textbf{Accuracy} & \textbf{F1 score} & \textbf{\makecell{Cohen's\\kappa}} \\
        \hline
        1&  GPT-4&100 \% & 94.7 \% & 0.94 & 0.88\\
        2&  GPT-4, Sonnet 3-5&94.5 \% & 96.3 \% & 0.96 & 0.92\\
        3&  GPT-4, Sonnet 3-5, Llama-3-70B&89.25 \% & 97.9 \% & 0.98& 0.95\\
        4 &  GPT-4, Sonnet 3-5, Llama-3-70B, GPT-4o&86.38 \% & 98.1 \% & 0.98 & 0.95\\
        \hline
    \end{tabular}
    \caption{Performance metrics of large language models in multi-model agreement scenarios. Accuracy, F1 score, and Cohen's kappa improve as more models agree on predictions.}
    \label{tab:model_metrics}
\end{table}

Table \ref{tab:model_metrics} shows how grading performance improves as we consider agreement between multiple models and in their assessment. The number of models in agreement indicates the models (in descending order of their performance) that agree in terms of their score on a particular transcript. The models are considered in descending order of their individual performance: GPT-4, Claude-Sonnet-3-5, Llama-3-70B, and GPT-4o. The "Number of models in agreement" column in Table 7 should be interpreted as follows \textemdash "1": Results from GPT-4 alone, "2": Instances where both GPT-4 and Claude-Sonnet-3-5 agree, 
"3": Instances where GPT-4, Claude-Sonnet-3-5, and Llama-3-70B all agree
"All 4 models": Instances where all four models (including GPT-4o) agree on the score. This hierarchical agreement approach allows us to observe how the inclusion of additional, slightly less accurate models in the consensus can impact overall grading performance. The results in Table \ref{tab:model_metrics} reveal a clear trend: as the number of agreeing models increases, we see a consistent improvement in all performance metrics (Accuracy, F1 score, and Cohen's kappa). This improvement comes at the cost of coverage of student exams, as the percentage of the dataset meeting the agreement criteria decreases. However, even at the most stringent level (all 4 models agreeing), we still cover 86.38 \% of the dataset, indicating a high level of consensus among these diverse models for a large portion of the grading decisions. As we move from requiring agreement from just one model to all four models, we see a consistent improvement across all performance metrics. Accuracy increases from 94.7\% to 98.1\%, F1 score from 0.94 to 0.98, and Cohen's kappa from 0.879 to 0.952. When all four models agree, we achieve the highest performance with 98.1\% accuracy and a Cohen's kappa of 0.952, indicating almost perfect alignment with human graders. This suggests that AI based grading may be most reliable when several models agree with each other.

Analysis of agreement rates across grades in Table \ref{tab:grade_agreement}  shows a significant difference in agreement rates when the awarded grade is 0 or 1 ($\chi^2 = 40.8715$, $p < 0.0001$). When the grade is 1 - the models exhibit higher agreement rate (89.41\%) when compared to when grade is 0 (77.47\%). This is further supported by the Fisher's exact test (OR = 0.4040, $p < 0.0001$), indicating that the odds of model agreement when grade is 1 is much higher than when the grade is 0. As seen in Table \ref{tab:year_agreement}), there is no such significant difference in model agreement across years.

Table \ref{tab:combined_results} presents the overall chi-square test results for whether the model agreement varies across the different stations. The results indicate a significant difference in agreement rates across medical stations ($\chi^2 = 17.0204$, $p = 0.0299$). Further, pairwise comparison between stations showed showed a significant difference in model agreement rates between memory problems station and the vaginal discharge station. There was no significant difference in the model agreement rate for the other stations.

\begin{table}[htbp]
\centering
\begin{tabular}{lrrr}
\hline
\textbf{Grade} & \textbf{Total N} & \textbf{Agreement Rate} & \textbf{Disagreement Rate} \\
\hline
0 & 506 & 77.47\% & 22.53\% \\
1 & 1510 & 89.41\% & 10.59\% \\
\hline
\multicolumn{4}{l}{\textit{Statistical Tests}} \\
\hline
\multicolumn{2}{l}{Chi-square test} & $\chi^2 = 40.8715$ & $p < 0.0001$ \\
\multicolumn{2}{l}{Fisher's exact test} & OR = 0.4040 & $p < 0.0001$ \\
\hline
\end{tabular}
\caption{Comparison of model agreement rates for different grades: 0 and 1.}
\label{tab:grade_agreement}
\end{table}

\begin{table}[htbp]
\centering
\begin{tabular}{lrrr}
\hline
\textbf{Year} & \textbf{Total N} & \textbf{Agreement Rate} & \textbf{Disagreement Rate} \\
\hline
2019 & 585 & 85.47\% & 14.53\% \\
2020 & 316 & 86.08\% & 13.92\% \\
2021 & 470 & 85.96\% & 14.04\% \\
2022 & 645 & 87.75\% & 12.25\% \\
\hline
\multicolumn{4}{l}{\textit{Statistical Tests}} \\
\hline
\multicolumn{2}{l}{Chi-square test} & $\chi^2 = 2.6843$ & $p = 0.4432$ \\
\multicolumn{2}{l}{Effect Size (Cramer's V)} & \multicolumn{2}{r}{0.0365} \\
\hline
\end{tabular}
\caption{Comparison of model agreement rates for different years: 2019, 2020, 2021, 2022.}
\label{tab:year_agreement}
\end{table}

\begin{table}[htbp]
\centering
\begin{tabular}{lr}
\multicolumn{2}{c}{\textbf{Chi-square test results for comparing model agreement across stations}} \\
\hline
\textbf{Statistic} & \textbf{Value} \\
\hline
Chi-square & 17.0204 \\
Degrees of freedom & 8 \\
p-value & 0.0299 \\
Cramer's V & 0.0919 \\
\hline
\end{tabular}

\vspace{0.5cm}

\begin{tabular}{llrrr}
\multicolumn{5}{c}{\textbf{Top most significant pairwise comparisons between different  medical stations in terms of model agreement}} \\
\hline
\textbf{Station 1} & \textbf{Station 2} & \textbf{Chi-square} & \textbf{p-value} & \textbf{Corrected p-value} \\
\hline
Memory problems & Vaginal discharge & 10.604 & 0.001 & 0.041* \\
Leg pain & Memory problems & 7.157 & 0.007 & 0.269 \\
Memory problems & Vision problems & 5.203 & 0.023 & 0.812 \\
\hline
\multicolumn{5}{l}{\small *Significant at p < 0.05 after Bonferroni correction} \\
\end{tabular}
\caption{Chi-square test results and most significant pairwise comparisons of agreement rates across medical stations}
\label{tab:combined_results}
\end{table}

\section{Discussion} \label{sec:disc}

\subsection{Key takeaways}
The frontier LLMs, particularly GPT-4, demonstrated exceptional alignment with human graders, achieving 95.7\% accuracy, 0.95 F1 score, and a Cohen's kappa of 0.88. Claude-Sonnet-3-5 and Llama-3-70B also showed strong performance with 93.6\% and 93.2\% accuracy respectively, and kappa values over 0.83. The less capable models GPT-3.5, Mixtral 8x7B and Starling -7B demonstrated much worse alignment with human graders, all reaching a Cohens' kappa of less than 0.35, underscoring the rapid pace of advances in LLM capabilities. Error analysis, described in greater detail below, showed that these poor-performing models consistently fail to understand what constitutes a summary of medical history and hence are easily fooled by other statements that may contain medical history but are statements that are either related to collecting medical history or providing a diagnosis to the patient.

Notably, the model-human alignment was poor for years 2019 and 2020 (kappa 0.42-0.61), but excellent  for 2021 and 2022 (kappa 0.80-0.88), indicating most likely a significant change in the human grading process for the latter two years. This drastic improvement hints at potential inefficiencies in the previous grading process at the medical school. In several specific instances of disagreement, we manually verified that the AI could detect the relevant portion of the transcript to support its score and provided a rationale, thus increasing confidence in its grading. The human grading mistakes revealed by applying AI to this process suggest that AI-assisted grading could potentially enhance the consistency and thoroughness of human assessment processes. Along these lines, ensemble methods show promise for further improving the quality and reliability of an LLM-based grading system. Our analysis on inter-model agreement revealed that consensus among multiple models strongly correlates with increased accuracy. When all four top models agreed, we achieved > 98\% accuracy and a Cohen's kappa > 0.95.  

We also investigated a retrieval augmented grading strategies and found that while achieving high recall rates (up to 92\% within the top 5 chunks), the RAG approach under-performed compared to full transcript grading. This performance gap highlighted the importance of full conversational context to accurately assess communication skills. 

\subsection{Failure analysis}
Our analysis also revealed limitations and potential failure modes that must be considered when implementing such a transcript-based  AI grading system. Addressing these challenges will be crucial in ensuring the reliability and fairness of AI-assisted assessment in medical education. The two main categories errors were centered around data quality and LLM inference.
\subsubsection*{Data errors.}
\begin{itemize}
    \item Audio recording errors: We found that we had to discard ~2\% of our transcripts due to poor quality caused by audio recording errors. This could be due to several reasons such as mics not being setup properly, recording not starting at the right time, or too much background noise muffling the conversation. In future deployments, before using a transcript for AI-assessment,  sufficient audio quality should be first verified for faithful transcription.
    \item Transcription errors: The Automatic Speech Recognition (ASR) system's quality significantly influences downstream analysis outcomes. Whisper-v3 demonstrated excellent performance, producing high-quality transcripts even from sub-optimal audio recordings. Primary error types included hallucinations during silent periods \textemdash a known limitation of transformer-based ASR models, and misspellings of medical and technical terminology \cite{radford2023robust}. However, these errors minimally impacted downstream analysis, as advanced Language Models (LLMs) can infer correct terms from misspellings.
\end{itemize}
\subsubsection*{LLM inference errors.}
The most common error patterns in LLM scoring of student transcripts, in order of increasing sophistication, are:
\begin{itemize}
    \item Hallucination errors. LLMs may generate non-existent summaries or falsely score students based on self-summarized symptoms. Starling-7B was most prone to this, with GPT-3.5 and Mixtral-8x7B also exhibiting such errors. While realistic, these can be easily identified through transcript verification.
    \item Prompt Misinterpretation: Models may misunderstand the concept of a medical history summary. When no clear summary exists, they might misidentify discussions of medical history or other symptoms as valid summaries, leading to false positive scores. GPT-3.5 and Mixtral-8x7B were particularly susceptible.
    \item Contextual Confusion: Despite understanding summary concepts, models may misinterpret context. For instance, they might mistake a student's diagnostic explanation during symptom recap for a valid medical history summary. Ideally, such summaries should conclude the information gathering phase, prompting patient confirmation and correction. 
\end{itemize}

LLM inference errors could be mitigated by using ensemble approaches, as we observed reduction in the error rate of when consensus between multiple distinct model outputs was indicated. A more holistic approach to improving the system could involve establishing a continuous feedback loop where human experts regularly review samples of AI-graded exams. Developing such a hybrid human-AI workflow, where the LLM acts as an initial grader and decision support tool with human experts reviewing and finalizing grades, could leverage the strengths of both automated and human assessment. Feedback from the human could also be utilized in a reinforcement learning manner to continously update model behavior to reflect changing grading strategies.

\subsection{Ethical considerations}
While our results demonstrate high alignment between LLM and human graders, as efforts are made to deploy AI assessment systems in practice, vigilance should be maintained for addressing ethical concerns.  On-going research as well as active monitoring will be necessary to ensure LLM-based systems do not exhibit unintended biases in performance assessment and maintain fair/equitable grading across diverse student populations. As more complex models emerge, preserving transparency in grade determination is crucial for student trust and feedback. The implementation of AI-based grading may influence OSCE preparation strategies; therefore, measures must be taken to ensure students develop genuine communication skills rather than optimizing for AI evaluation. Proactively addressing these ethical considerations will enable effective leveraging of AI benefits in medical education assessment while mitigating potential risks and ensuring fairness and transparency.

We plan to extend this work to all sections of the communication rubric \textemdash assessing student ability in  providing information, gathering information, showing empathy and decision making. Additionally, further research may be focused on developing standardized guidelines for the implementation of AI-based grading systems. Investigating the applicability of these techniques to other aspects of medical education, such as curriculum development and personalized learning, could lead to a more comprehensive integration of AI in medical training.

\section{Conclusion} \label{sec:conc}

To our knowledge, this is the first work to study use of  LLMs towards analyzing medical student OSCE transcripts for automatic assessment. Given the relatively large sample size of our study (with over 2000 OSCE encounter transcripts over 10 different stations) we believe our these findings provide a solid foundation for better understanding the strengths and limitations of GenAI-based grading in medical education with currently available AI models. 


\bibliographystyle{unsrt}  
\bibliography{references}  

\newpage
\section*{Supplementary Material}
\setcounter{subsection}{0}
\subsection*{1. Prompts} \label{sec:prompts}

\begin{tcolorbox}[examplebox={Prompt: Summary of history grader}]
\begin{lstlisting}
You are an AI assistant tasked with analyzing a transcript of a medical encounter between a medical student and a patient actor.
Your task is to:
  a) detect whether the student summarized the patient's medical history
  b) extract the exact statements used by the student to summarize the medical history. 

SUMMARIZATION STATEMENT
- The statement should be in the provided transcript of the dialogue.
- The statement should be spoken by the medical student.
- The statements must summarize the medical history provided by the patient
- The statements should repeat or rephrase the key information about the patients medical history collected from the patient.
- The statements may begin with phrases like 'to sum up', 'to reiterate', 'in summary', or 'to summarize' etc. 
- The statements SHOULD NOT include the questions asked to collect information from the patient.
- The statements should be DISTINCT from any discussion about the diagnosis of the patients condition.
- The statements should be DISTINCT from any discussion about the next steps to be taken by the patient. 

This is the excerpt of the transcript of the conversation between the medical student and the patient within the placeholders '<>': <{context}>

INSTRUCTIONS
- Identify whether the student summarized the patients medical history.
- Extract the exact statement used by the student to summarize the patients medical history
- If the student did not summarize the patients medical history, then respond with "summary not found in transcript"
- Assign a score to the student based on the statement that were extracted
- Remember that the summary is DISTINCT from any discussion about the diagnosis or next steps.
- Return your response in JSON format

SCORING
Give the student a score of,
0: The student failed to provide a distinct summary of the patients medical history.
1: The student provided an adequate and distinct summary of the patients medical history.

Give the student at score of 1 only if
- The statement is not a question or an enquiry about symtoms.
- The statement is spoken by the medical student.
- The statement shows that the student summarized the medical history reported by the patient
- The statement is DISTINCT from any discussion about the diagnosis of the patients condition.
- The statement is DISTINCT from any discussion about the next steps to be taken by the patient. 

Remember that the statement should be in the provided transcript of the dialogue.

ANSWER FORMAT
Return your response in the following JSON format:
{{
"statement": "exact statement extracted from the transcript",
"rationale": "explanation of the students score based on the statement, who spoke the statement and the SCORING instructions",
"score": 0 / 1
}}
\end{lstlisting}
\end{tcolorbox}

\subsection*{2. Prompts to generate queries for RAG} \label{sec:rag_prompts}

\begin{tcolorbox}[examplebox={Prompt: hypothetical document embedding query generation - HyDE}]
\begin{lstlisting}
Imagine you are a medical student examining a patient actor in a simulated medical exam.
The patient presents you with a condition of {case} and describes their symptoms and medical history.
Your task is to listen to the patient and tell them a summary of their symptoms and medical history. Just give them a summary and nothing else. What would you say?
\end{lstlisting}
\end{tcolorbox}

\begin{tcolorbox}[examplebox={Prompt: auto-summarization query generation}]
\begin{lstlisting}
You are an AI assistant tasked with analyzing excerpts of a conversation between students and patient actors in a simulated medical exam. 
You will be provided with the transcript of one such exam. You need to read through the transcript and extract significant pieces of information.
For this task,
- you need to find the medical history being reported by the patient during the conversation.
- you need to summarise the patient's medical history.
- The summary that you return should be as if the student is summarizing what they have heard back to the patient.
- The summary that you return should be as if it is being spoken back to the patient.

This is the transcript of the conversation between student and medical history: {context}

Remember that:
- All the information that you return must be found in the transcript that was provided to you. 
- Do not include any other information in your response except for the medical history of the patient.
- Do not include the diagnosis that was made by the medical student
- Do not include the next steps or workup that were proposed by the medical student

Your response should be a 2 -3 line summary of the patient's medical history and symptoms. Your reponse should be as if it was are spoken by the medical student back to the patient. 
\end{lstlisting}
\end{tcolorbox}

\subsection*{3. Examples of LLM errors and failure modes}

i) An example of a failure mode where the model hallucinates, i.e., makes up a nearly perfect summary statement that does not actually exist in the transcript. 
\begin{tcolorbox}[examplebox={Example: hallucination errors}]
\begin{lstlisting}x
{
  "statement": "So just to kind of recap and to make sure that I understood everything. So you came in today because of memory problems that started seven months ago. And you've been feeling that you can't concentrate or stay focused. But a month before all this stuff started happening, your wife had passed away. And so you're going to be in a lot of trouble. You're feeling sad and lonely. You haven't been able to concentrate and stay in focus. You have lost your appetite. You can not sleep. You also have not lost interest in your hobbies. And anything else in terms of what's been going on, or is that pretty much covered?",
  "rationale": "The student provided a distinct and adequate summary of the patient's medical history, repeating and rephrasing key information collected from the patient. This summary is distinct from any discussion about the diagnosis or next steps, focusing solely on the patient's reported symptoms and history. The student's statement is not a question but a recap, fulfilling the criteria for a score of 1",
  "score": 1
}
\end{lstlisting}
\end{tcolorbox}

ii) An example of an error where the model misunderstands the task at hand. Here, the model has misinterpreted the act of collecting medical history as providing a summary. The given statement is a series of questions and responses between the doctor and patient, not a summary.

\begin{tcolorbox}[examplebox={Example: misunderstanding errors}]
\begin{lstlisting}
{
  "statement": "I've been having this pain in my leg. Oh, when i was walking it when I was started walking okay while you're walking okay do you normally walk or was that just kind of have you been kind of doing more activity recently? No, I walk on a daily basis. Okay and it's while you were walking and is it, and you pointed to your left, is that on your website? Yes, it is. Okay just on your life? Mm-hmm. Okay nothing on your right? No. Okay all right and so how far do you walk?",
  "rationale": "The student summarized the patient's medical history by asking about the onset of the pain, the location of the pain, and the patient's daily activity. The student also confirmed the patient's medical history of high blood pressure and diabetes. The statement is spoken by the medical student and is distinct from any discussion about the diagnosis or next steps. Therefore, the student receives a score of 1.",
  "score": 1
}
\end{lstlisting}
\end{tcolorbox}

iii) An example of an error where the model mistakes the context at which the summary has been provided. While the LLM correctly identified a summary, it failed to recognize that this summary was given in the context of explaining a diagnosis, not as a standalone summary of the patient's medical history. A proper summary should be given earlier in the conversation, after collecting information and before moving to diagnosis or treatment plans.
\begin{tcolorbox}[examplebox={Example: contextual confusion errors}]
\begin{lstlisting}
{
  "statement": "Based on your symptoms, and the fact that you've been experiencing a persistent cough for two weeks, along with chest discomfort and fatigue, I think you may have bronchitis",
  "rationale": "The student provided an excellent summary of the patient's medical history by recapping the symptoms: persistent cough for two weeks, chest discomfort, and fatigue.",
  "score": 1
}
\end{lstlisting}
\end{tcolorbox}
\end{document}